\documentclass{article}

\usepackage{graphicx}
\usepackage{wrapfig}
\usepackage{enumitem}
\usepackage{algorithm}
\usepackage{algcompatible}
\usepackage{wrapfig}

\usepackage[final]{corl_2020} 

\title{COG: Connecting New Skills to Past Experience\\ with Offline Reinforcement Learning}

\author{
  Avi Singh, Albert Yu, Jonathan Yang, Jesse Zhang, Aviral Kumar, Sergey Levine\\
  University of California, Berkeley \\ 
  \texttt{\{avisingh,albertyu,jy2370,jessezhang,aviralk,svlevine\}@berkeley.edu} \\
}

\usepackage{amsfonts}
\usepackage{amsmath}
\usepackage{amsthm}
\usepackage{tcolorbox}
\usepackage{setspace}

\newcommand{\policy}{\pi}

\newcommand{\transitions}{T}

\newcommand{\bellman}{\mathcal{B}}

\newcommand{\bs}{\mathbf{s}}
\newcommand{\ba}{\mathbf{a}}

\newcommand{\E}{\mathbb{E}}

\begin{document}
\maketitle

\begin{abstract}
Reinforcement learning has been applied to a wide variety of robotics problems, but most of such applications involve collecting data from scratch for each new task. Since the amount of robot data we can collect for any single task is limited by time and cost considerations, the learned behavior is typically narrow: the policy can only execute the task in a handful of scenarios that it was trained on. What if there was a way to incorporate a large amount of prior data, either from previously solved tasks or from unsupervised or undirected environment interaction, to extend and generalize learned behaviors? 
While most prior work on extending robotic skills using pre-collected data focuses on building explicit hierarchies or skill decompositions, we show in this paper that we can reuse prior data to extend new skills simply through dynamic programming.
We show that even when the prior data does not actually succeed at solving the new task, it can still be utilized for learning a better policy, by providing the agent with a broader understanding of the mechanics of its environment.
We demonstrate the effectiveness of our approach by chaining together several behaviors seen in prior datasets for solving a new task, with our hardest experimental setting involving composing four robotic skills in a row: picking, placing, drawer opening, and grasping, where a +1/0 sparse reward is provided only on task completion. We train our policies in an end-to-end fashion, mapping high-dimensional image observations to low-level robot control commands, and present results in both simulated and real world domains. 
Additional materials and source code can be found on our project website: \url{https://sites.google.com/view/cog-rl}. 

\end{abstract}


\vspace{-4pt}
\section{Introduction}
\label{sec:intro}
\vspace{-5pt}

Consider a robot that has been trained using reinforcement learning (RL) to take an object out of an open drawer. It learns to grasp the object and pull it out of the drawer. If the robot is then placed in a scene where the drawer is instead closed, it will likely fail to take the object out, since it has not seen this {scenario or} initial condition before. How can we enable learning-based robotic systems to reason effectively about such scenarios?  
Conventionally, we might expect that complex methods based on hierarchies or explicit skill decomposition would be needed to integrate a drawer opening skill with the grasping behavior. But what if simply combining previously collected (and \textit{unlabeled}) robot interaction data, which might include drawer opening and other behaviors,  together with offline RL methods~\citep{levine2020offline}, can allow these behaviors to be combined \emph{automatically}, without any explicit separation into individual skills? In this paper, we study how model-free RL algorithms can utilize prior data to extend and generalize learned behaviors, incorporating segments of experience from this prior data as needed at test-time.

\begin{figure}[t]
\begin{center}
    \vspace{-0.1cm}
      \includegraphics[width=0.97\linewidth]{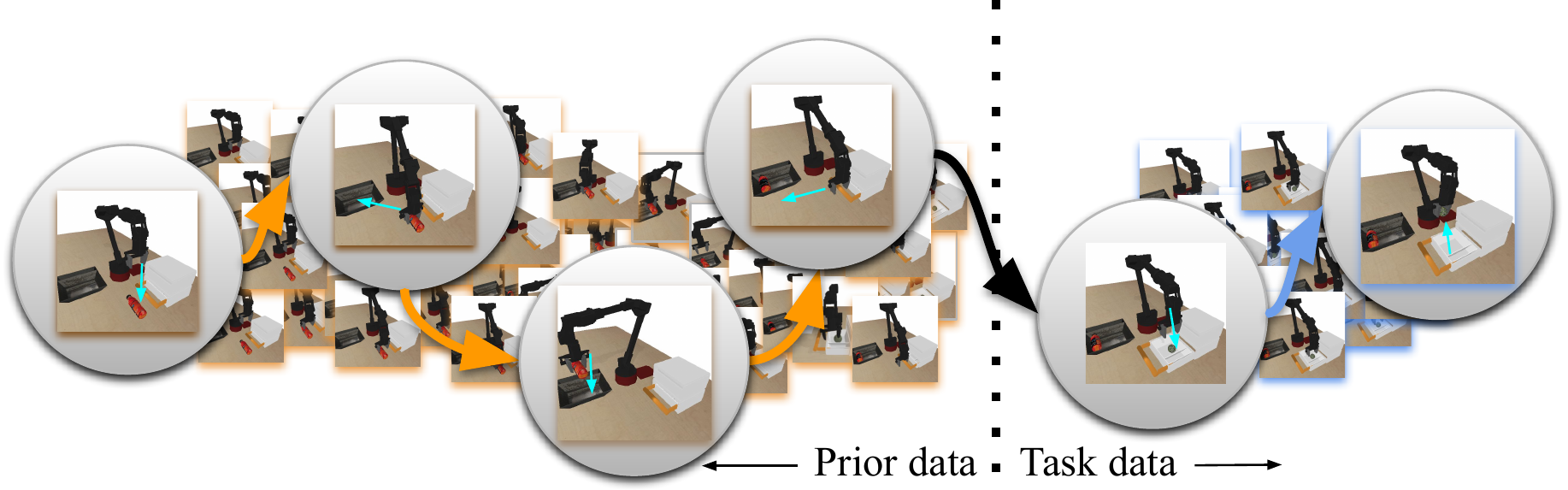}
    \vspace{-7pt}
    \caption{\footnotesize{\textbf{Incorporating unlabeled prior data into the process of learning a new skill.} {We present a system that allows us to extend and generalize robotic skills by using unlabeled prior datasets. Learning a new skill requires collecting some task-specific data (right), which may not contain all the necessary behaviors needed to set up the initial conditions for this skill in a new setting (e.g., opening a drawer before taking something out of it). The prior data (left) can be used by the robot to automatically figure out that, when it encounters a closed drawer at test time, it can first open it, and then remove the object. The task data does not contain drawer opening, and the prior data does not contain any examples of lifting the new object.}}}
    \label{fig:chain-behavior-orig}
\end{center}
 \vspace{-0.7cm}
\end{figure}

Standard online RL algorithms typically require a tight interaction loop between data collection and policy learning. 
There has been a significant amount of recent interest in devising methods for offline RL~\citep{levine2020offline, fujimoto2018off, kumar2020conservative, kumar2019stabilizing, agarwal2019optimistic,jaques2019way}, which can leverage previously collected datasets without environment interaction. In this paper, we build on recent advances in offline RL, and show that a ``data-driven'' RL paradigm allows us to build real-world robotic learning systems that can learn increasingly flexible and general skills.
We will show that collecting a large dataset that covers a large repertoire of skills provides a useful foundation for learning new tasks, enabling learned policies to perform multi-stage tasks from previously unseen conditions, despite never having seen all of the stages together in a single episode. Consider the drawer example from before: if the robot can draw on its prior experience of drawer opening, it can take an object out of the drawer even if the drawer is initially closed, by figuring out that opening the drawer will place it into a state from which the grasping skill can succeed. Furthermore, even if there is an obstruction in front of the drawer, the robot can once again draw on the prior data, which might also include pick and place behaviors, to remove the obstruction. Crucially, this does not require the robot to practice drawer opening again when learning a new task -- it can collect data for the new task in isolation, and combine it with past data to compose novel behaviors. 
We illustrate this idea in Figure~\ref{fig:chain-behavior-orig}.

How can this prior data from different tasks be utilized for learning a new task? If the prior data contains successful behavior for the new task, this would be easy. However, we will show that even prior data that is unsuccessful at the new task can be useful for solving it, since it provides the policy with a deeper understanding of the mechanics of the world. In contrast to prior works that approach this problem from a hierarchical skill learning~\citep{fazeli2019see,nair2019hierarchical} or planning perspective~\citep{Ebert18journal, Eysenbach19Search}, we show that this ``understanding'' can be acquired without any explicit predictive model or planning, simply via model-free RL. 
Our method builds on a recently proposed model-free offline RL method called conservative Q-learning (CQL)~\cite{kumar2020conservative}.
We extend CQL to our problem setting as following: we initialize the replay buffer with our prior dataset, and assign zero reward to every transition in this dataset, since the new task is different from the prior executed behavior. We then run CQL on both this prior dataset and data collected specifically for the new task, and evaluate the policy on a variety of {initial conditions} which were not seen in the task-specific data. {We can further fine-tune the policy resulting from offline training with limited online data collection, which further improves the performance of the new skill while retaining the aforementioned generalization benefits.}  

Our main contribution is to demonstrate that model-free offline RL can learn to combine task data with prior data, producing previously unseen combinations of skills to meet the pre-conditions of the task of interest, and effectively generalizing to new initial conditions.
We call our approach \textbf{COG}: Connecting skills via Offline RL for Generalization.
We describe a robotic learning system that builds on this idea to learn a variety of manipulation behaviors, directly from images.
We evaluate our system on several manipulation tasks, where we collect prior datasets consisting of imperfect scripted behaviors, such as grasping, pulling, pushing, and placing a variety of objects. We use this prior data to learn several downstream skills: opening and closing drawers, taking objects out of drawers, putting objects in a tray, and so on. We train neural network-based policies on raw, high-dimensional image observations, and only use sparse binary rewards as supervision. We demonstrate the effectiveness of COG in both simulated domains and on a real world low-cost robotic arm.

\section{Related Work}
\vspace{-5pt}
\textbf{Robotic RL.}
RL has been applied to a wide variety of robotic manipulation tasks, including grasping objects~\cite{qt-opt, zeng2018learning}, in-hand object manipulation~\cite{dexterity, VanHoofInHandManipulation, dexterous1, dexterous2}, pouring fluids~\cite{schenck2017visual}, door opening~\cite{yahya2017collective, dooropening}, and manipulating cloth~\cite{matas18, vice-raq}. 
Most of these works use online RL methods, relying on a well-tuned interaction loop between data collection and policy training, instead of leveraging prior datasets. Our paper is more closely related to \citet{qt-opt},  \citet{julian2020efficient}, and \citet{cabi2019}, which also use offline RL and large prior datasets. However, these prior works focus on generalization to new objects, as well as fine-tuning to handle greater variability (e.g., more object types, changes in lighting, etc.). Our work instead focuses on changes in initial conditions that require entirely different skills than those learned as part of the current task, such as opening a drawer before grasping an object.

\textbf{Data-driven robotic learning.}
In addition to RL-based robotics, data-driven robotics in general has become increasingly popular in recent years, and several works have investigated using large-scale datasets to tackle long-standing challenges in robotics, such as grasping novel objects. 
However, most prior work in this category focuses on executing the same actions on novel objects~\citep{kappler2015bigdada, pinto2016supersizing, Levine16googlegrasping, Mahler16Dexnet1, gupta2018robot}. In contrast, we explicitly target problems where new behavior needs to be learned to perform the task in a new scenario, and use prior interaction datasets to achieve this ability via model-free RL.
Visual foresight~\cite{finn2017deep} and its followups~\cite{ebert2018visual, Ebert18journal,xie2019improvisation,hristov2018interpretable} also address temporally extended tasks with large datasets by learning video prediction models, but with significantly shorter time horizons than demonstrated in our work, due to the difficulty of long-horizon video prediction. \citet{mandlekar2019iris} use an alternate approach of explicitly learning hierarchical policies for control, and \citet{m2020learning} utilizes offline imitation learning to compose different demonstration trajectories together.

\textbf{Offline deep RL.} While offline (or ``batch'') RL is a well-studied area~\citep{lange2012,lagoudakis2003least,munos2005error,levine2020offline}, there has been a significant amount of recent interest in offline deep RL, where deep RL agents are trained on a static, previously collected dataset without any environment interaction~\citep{levine2020offline,kumar2020conservative,fujimoto2018off,kumar2019stabilizing,agarwal2019optimistic,wu2019behavior,peng2019awr,yu2020mopo}.
These works largely focuses on developing more effective offline deep RL algorithms by correcting for distribution shift~\citep{fu2019diagnosing,kumar2020conservative,kumar2019stabilizing,wu2019behavior,peng2019awr,yu2020mopo} which renders standard off-policy RL algorithms inadmissible in purely offline settings~\citep{kumar2019stabilizing,levine2020offline}. In contrast, our work does not propose a new algorithm, but rather adapts existing offline RL methods to the setting where prior data from a \emph{different} domain must be integrated into learning a new task such that it succeeds under a variety of conditions.

\newcommand{\task}[0]{\mathbb{T}}
\newcommand{\dataprior}[0]{\mathcal{D}_\text{prior}}
\newcommand{\datatask}[0]{\mathcal{D}_\task}

\vspace{-7pt}
\section{Incorporating Prior Data into Robotic Reinforcement Learning}
\label{sec:problem-statement}
\vspace{-7pt}

Our goal is to develop a system that can learn new robotic skills with reinforcement learning, while incorporating diverse previously collected data to provide greater generalization. We hypothesize that, {even} if a \textit{subset of} this prior data illustrates useful behaviors that are \emph{in support} of the new skill, including this data into the training process via offline RL can endow the robot with the {ability to reason}, such that when the conditions at test-time do not match those seen in the data for the specific new skill, the robot might still generalize because other transitions in the prior data have allowed it to learn how to react intelligently.
For instance, when a robot trained to perform object grasping is faced with a closed or obstructed drawer at test time, without any prior experience, the robot would have no way to know how to react. Humans generally handle such situations gracefully, by drawing on their past experience. We use the term ``common sense'' to refer to this ability to handle small test-time variations in the task, such as the need to open a drawer before taking something out of it. How can we endow learned policies with this sort of ``common sense''?
If the prior training data illustrates opening of drawers and removing obstructions, then even without being told that such behaviors are useful for grasping an object from an obstructed drawer, model-free offline RL methods can learn to reason and utilize such behaviors at test time. Crucially, we will show how these ``common sense'' behaviors emerge entirely from combining rich prior data and offline RL, without any explicit reasoning about preconditions or hierarchical higher-level planning.

We formalize our problem in the standard RL framework. The goal in RL is to optimize the infinite horizon discounted return  $R_t = \sum_{t = 0}^{\infty}\gamma^t r_t$ in a Markov decision process (MDP), which is defined by a tuple $(\mathcal{S}, \mathcal{A}, \transitions, r, \gamma)$, where
$\mathcal{S}$ and $\mathcal{A}$ represent state and action spaces, $\transitions(s' | s, a)$ and $r(s,a)$ represent the dynamics and reward function, and $\gamma \in (0,1)$ represents the discount factor. We operate in the \emph{offline} RL setting as opposed to the standard online regime since we are interested in leveraging most out of prior datasets. 

In most prior works on offline RL~\citep{kumar2019stabilizing,fujimoto2018off,agarwal2019optimistic}, the method is typically provided with data for the specific task that we wish to train a policy for, and the entire dataset is annotated with rewards that defines our objective for that task. In contrast, there are two distinct sources of data in our problem setting: task-agnostic, unlabeled prior data, which we denote as $\dataprior$, and task-specific data, which we denote as $\datatask$, where $\task$ represents our task. The datapoints in $\dataprior$ simply consist of $(s, a, s')$ transitions, and do not have any associated reward labels. While we cannot train a policy to achieve any particular objective from this data alone, it is informative about the dynamics of the MDP where the data was collected. On the other hand, the datapoints in $\datatask$ consist of $(s, a, s', r)$ tuples, and can be used for learning a policy that maximizes the observed reward. To summarize, the input and output of our problem setting are as follows: 
\begin{tcolorbox}
\textbf{Input: } Datasets $\dataprior$ (with no reward annotations), $\datatask$ (with sparse rewards for task $\task$). 

\textbf{Return: } Policy $\pi$ trained to execute task $\task$, which should be able to generalize broadly to new initial conditions. We would like to leverage $\dataprior$ for the latter. 
\end{tcolorbox}

\vspace{-5pt}
\section{Connecting New Skills to Past Experience via Dynamic Programming}
\label{sec:method-main}
\vspace{-5pt}

Our approach is conceptually very simple: use offline RL to incorporate prior data into the training for the new skill. However, the reasons why this approach should be effective in our problem setting are somewhat nuanced. In this section, we will discuss how model-free dynamic programming methods based on Q-learning can be used to connect new tasks to past experience. Before presenting COG, let's first briefly revisit off-policy deep RL algorithms. 

\begin{figure}[t]
\begin{center}
    \includegraphics[width=0.95\linewidth]{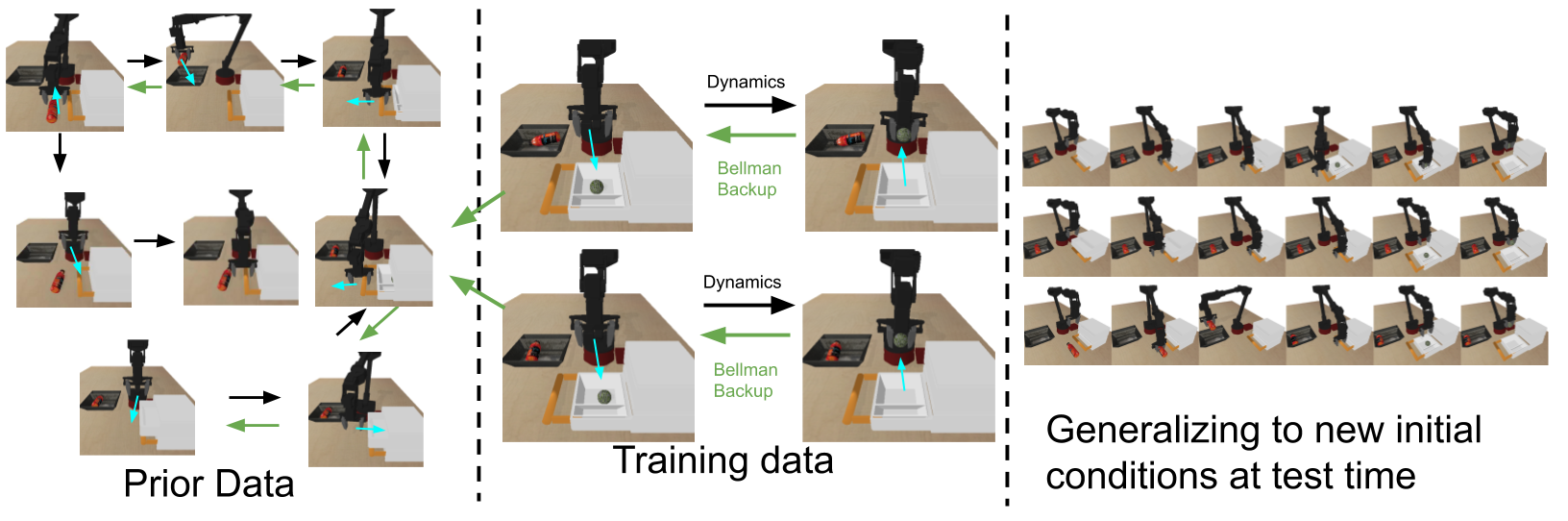}
    \vspace{-10pt}
    \caption{\footnotesize \textbf{Connecting new skills to past experience.} {Q-learning propagates information backwards in a trajectory (middle) and by stitching together trajectories via Bellman backups from the task-agnostic prior data (left), it can learn optimal actions from initial conditions appearing in the prior data (right).}}
    \label{fig:chain-behavior}
    \vspace{-0.5cm}
\end{center}
\end{figure}
{Standard off-policy deep RL methods}, such as SAC~\citep{haarnoja2018sacapps}, maintain a parametric action-value or Q-function, $Q_\theta(\bs, \ba)$,  and optionally a parametric policy, $\pi_\phi(\ba|\bs)$, with parameters $\theta$ and $\phi$, respectively. These methods typically train $Q_\theta(\bs, \ba)$ to predict the return under $\policy_\phi$ in the policy evaluation step, and then update $\policy_\phi$ in the direction of increasing Q-values in the policy improvement step:
\begin{small}
\begin{align*}
    \theta^{k+1} \leftarrow& \arg\min_{\theta} \E_{\bs, \ba, \bs' \sim \datatask}\left[ \left((r(\bs, \ba) + \gamma \E_{\ba' \sim \hat{\policy}^k(\ba'|\bs')}[\hat{Q}^{k}(\bs', \ba')]) - Q_\theta(\bs, \ba)\right)^2 \right]~\text{(policy evaluation)}\\ 
    \phi^{k+1} \leftarrow& \arg\max_{\phi} \E_{\bs \sim \datatask, \ba \sim \policy^{k}_{\phi}(\ba|\bs)}\left[\hat{Q}_\theta^{k+1}(\bs, \ba)\right]~~~ \text{(policy improvement)} 
\end{align*}
\end{small}
In order to see how this model-free procedure can help us stitch together different behaviors (which we will define shortly),
we start with some intuition about the Q-learning process.
Q-learning propagates information backwards through a trajectory. State-action pairs at the end of a trajectory with a high reward are assigned higher values, and these values propagate to states that are further back in time, all the way to the beginning of the trajectory. Recall the example from Section~\ref{sec:intro}, where the goal is to take an object out of a drawer. Assume that a reward of +1 is obtained when the object has been taken out of the drawer, and otherwise the reward is 0.
Under this reward, the state where the object has been taken out will have the highest value, and the state where the robot has grasped the object (but not yet lifted it) will have a slightly lower value, since it is farther away from the successful completion of the task. The initial state, where the robot gripper is far away from the object, will have very low value. See Figure~\ref{fig:chain-behavior} (right) for an illustration of such states. However, the initial state will still have a non-zero value, since the Q-function ``understands'' that it is possible to reach high-reward states. Any state for which there does not exist a valid path to a high-reward state will have a value that is equal to zero, and all other states will have non-zero values, decreasing exponentially (in the discount) with distance to the state where object is out of the drawer.

We may now ask, what happens if at test-time, the robot is asked to perform the task from some new state that was not seen in $\datatask$, such as a state where the drawer is closed? Such a state would either have a value of zero or, even worse, an arbitrary value, since it is outside of the training distribution. Therefore, the robot would {likely} not succeed at the task from this situation. Of course, we could train the new skill from a wider range of initial states, but if each skill must be learned from every possible starting state, it will quickly become prohibitively costly to train large skill repertoires.

{What if the Q-learning method was now augmented with an additional large dataset that contains a wide variety of other behaviors, but \emph{does not contain any data for the new task}? 
Such a dataset can still help us learn a much more useful policy, even in the absence of reward annotation: it can provide us with trajectories that (approximately) connect states not observed in $\datatask$ (e.g., a closed drawer) to states appearing in successful executions in $\datatask$ (e.g., open drawer), as shown in Figure~\ref{fig:chain-behavior}. If the prior dataset $\dataprior$ is large enough, it can inform the policy of different ways \emph{of reaching states from which the new task is solvable}. For example, if an object obstructs the drawer, and the prior dataset contains pick and place trajectories, then the policy can reason that it can unobstruct the drawer before opening it. Model-free Q-learning alone, without any skill decomposition or planning, can propagate values from $\datatask$ into $\dataprior$, allowing us to learn a policy that can execute the task from a much broader distribution of initial states without actually seeing full executions of the task from these states. Even without a single trajectory that both opens the drawer and takes the object out, as long as there is a non-zero overlap between $\dataprior$ and $\datatask$, Q-learning can still learn from $\dataprior$.}

\textbf{Offline RL via conservative Q-learning (CQL).} In order to incorporate the prior data $\dataprior$ into the RL process, we require an algorithm that can effectively utilize such prior data without actually interacting with the environment from the same initial states. Standard off-policy Q-learning and actor-critic algorithms are susceptible to out-of-distribution actions in this setting~\citep{levine2020offline,kumar2019stabilizing,kumar_blog}. We instead utilize the conservative Q-learning (CQL)~\citep{kumar2020conservative} algorithm that additionally penalizes Q-values on out-of-distribution actions during training. CQL learns Q-functions, ${Q}_\theta(\bs, \ba)$, such that the expected policy value under ${Q}_\theta$ lower-bounds the true policy value $\policy_\phi$, by minimizing the log-sum-exp of the Q-values at each state $\bs$, while maximizing the expected Q-value on the dataset action distribution, in addition to standard Bellman error training as shown in Equation~\ref{eqn:cql_h}. 
The training objective shown in Equation~\ref{eqn:cql_h} is the variant of CQL used in this paper:
\begin{equation}
    \small{\min_{Q}~ \alpha \mathbb{E}_{\bs \sim \dataprior \cup \datatask}\left[\log \sum_{\ba} \exp(Q(\bs, \ba))-\E_{\ba \sim \dataprior \cup \datatask}\left[Q(\bs, \ba)\right]\right] + \frac{1}{2} \E_{\bs, \ba, \bs' \sim \datatask \cup \dataprior}\left[\left(Q - {\bellman}^{\policy_k} \bar{Q} \right)^2 \right]\!.}
    \label{eqn:cql_h}
\end{equation}
We instantiate CQL as an actor-critic algorithm. The policy improvement step remains unchanged as compared to a standard off-policy RL method, as discussed previously.

{\textbf{Online fine-tuning.} In addition to the completely offline phase that utilizes the conservative Q-learning algorithm, we can further fine-tune the resulting policy using a small amount of online interaction with the environment. In order to do so, we train completely offline using CQL for 1m gradient steps, and then perform online training by periodically unrolling the policy in the environment and training on this additional data, starting from the solution obtained in the purely offline phase. We still utilize the CQL algorithm in this phase, except we only utilize the data collected via online interaction for fine-tuning purposes.}

\section{End-to-End Robotic Learning with Prior Datasets}
\label{sec:system-details}

In this section, we discuss how our method can be instantiated in a practical robotic learning system. 

\textbf{MDP for robotic skills.}
The state observation $s \in \mathcal{S}$ consists of the robot's current camera observation, which is an RGB image of size 48 $\times$ 48 for simulated experiments, and 64 $\times$ 64 for real robot experiments, and the current robot state. The robot state consists of the end-effector pose (Cartesian coordinates and Euler angles), and the extent to which the gripper is open (represented using a continuous value).
The action space $\mathcal{A}$ consists of six continuous actions and two discrete actions. The six continuous actions correspond to controlling the end-effector's 3D coordinates and its orientation. 
The first discrete action corresponds to opening or closing the gripper, while the second discrete action executes a return to the robot's starting configuration.
We use sparse reward functions for all of our tasks: a reward of +1 is provided when a task has been executed successfully, while a zero reward is provided for all other states. We do not have any terminal states in our MDP. 

\textbf{Data collection.} Our prior data collection takes place before any task-specific learning has happened. Since a completely random policy will seldom execute behaviors of interest, we bias our data collection policy towards executing more interesting behavior through the use of weak scripted policies: these policies typically have a success rate of 30-50\% depending on the complexity of the task they are performing. More details on the scripted policies can be found in Appendix~\ref{app:data_collection}.

\textbf{Neural network architectures.}
Since we learn to stitch together behaviors directly from raw, visual observation inputs, we utilize convolutional neural networks (ConvNets) to represent our policy $\pi_\phi$ and the Q-function $Q_\theta$. For both simulated and real world experiments, we use a 8-layer network: the first five layers consist of alternating convolutional and max-pooling layers (starting with a convolution), with a stride of 1 and kernel size of 3 for convolutional layers, and a stride of two for max-pooling layers. Each convolutional layer has 16 filters. We then pass the flattened output of the convolutional layer to three fully connected layers, of size 1024, 512 and 256. We use the ReLU non-linearity for all convolutional and fully-connected layers. A pictorial block diagram of these architectures is provided in Appendix~\ref{app:neural_net_arch}.

\vspace{-7pt}
\section{Experiments}
\label{sec:experiments}
\vspace{-10pt}

We aim to answer the following questions through our experiments: \textbf{(1)} Can model-free RL algorithms effectively leverage prior, task-agnostic robotic datasets for learning new skills? \textbf{(2)} Can our learned policies solve new tasks, especially from novel initial conditions, by stitching together behavior observed during training? \textbf{(3)} How does our approach compare to alternative methods for incorporating prior data (such as behavior cloning)? \textbf{(4)} Is the addition of prior data essential for learning to solve the new task? To this end, we evaluate our approach on a number of long-horizon, multi-step reasoning robotic tasks with different choices of $\datatask$ and $\dataprior$ and then perform an ablation study to understand the benefits of incorporating unlabeled offline datasets into robotic learning systems via offline reinforcement learning methods.

\vspace{-5pt}
\subsection{Experimental Setup}

\begin{wrapfigure}{r}{0.5\textwidth}
\begin{center}
  \vspace{-24pt}
  \includegraphics[width=\linewidth]{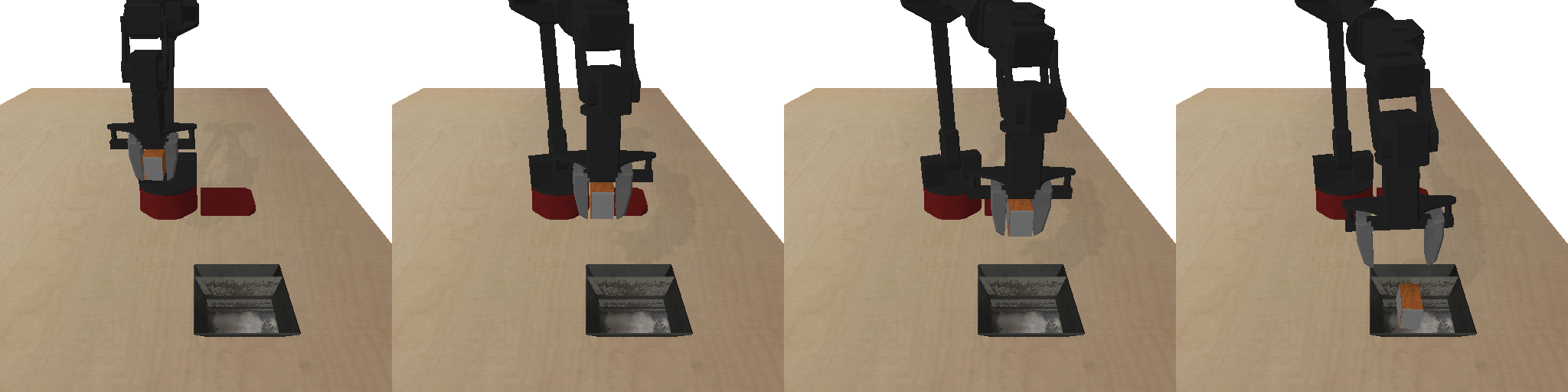}
  \includegraphics[width=\linewidth]{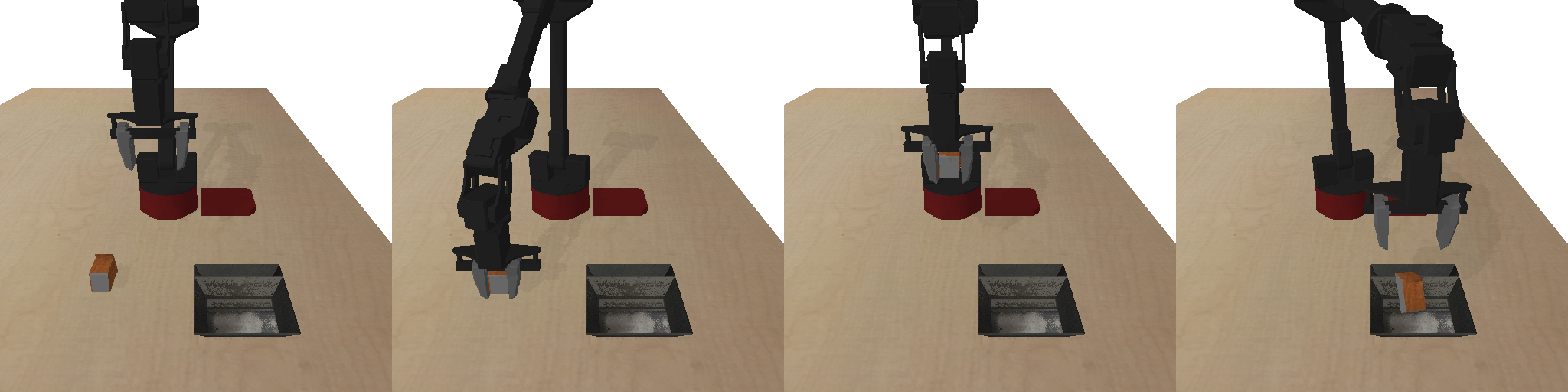}
      \caption{\footnotesize{\textbf{Picking and placing}. Example executions from our \textit{learned} policy. The first row shows the training condition, where the robot starts out already holding the object, and it only needs to place it in the tray. In the second condition (shown in second row), the  robot must first grasp the object before placing it into the tray.}}
     \label{fig:sim-task-pick-place}
     \vspace{-15pt}
 \end{center}
\end{wrapfigure}

\begin{wrapfigure}{r}{0.6\textwidth}
\small \begin{center}
\vspace{-25pt}
    \includegraphics[width=\linewidth]{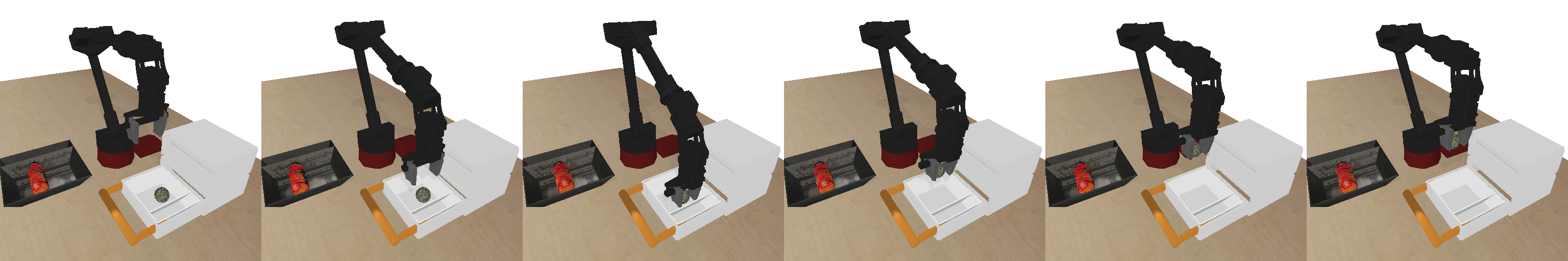}
    \includegraphics[width=\linewidth]{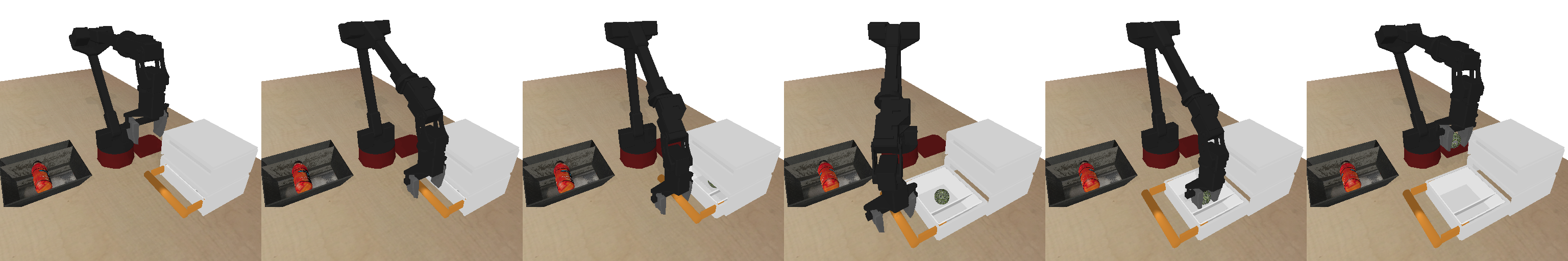}
    \includegraphics[width=\linewidth]{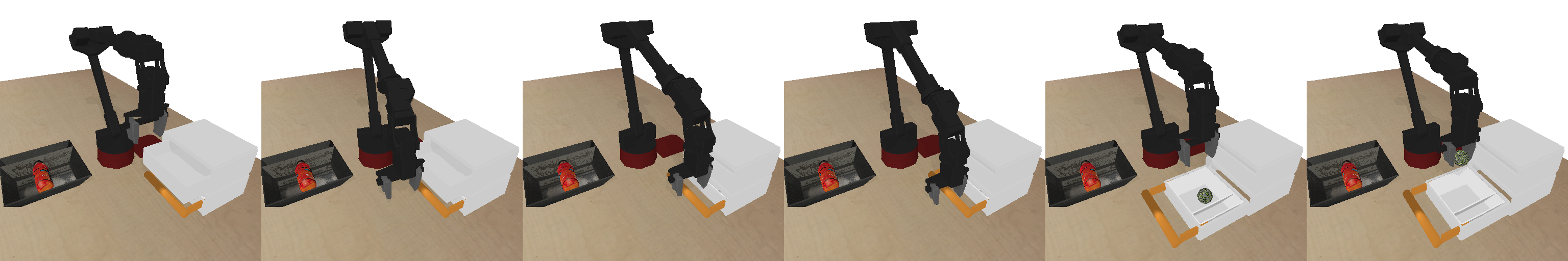}
    \includegraphics[width=\linewidth]{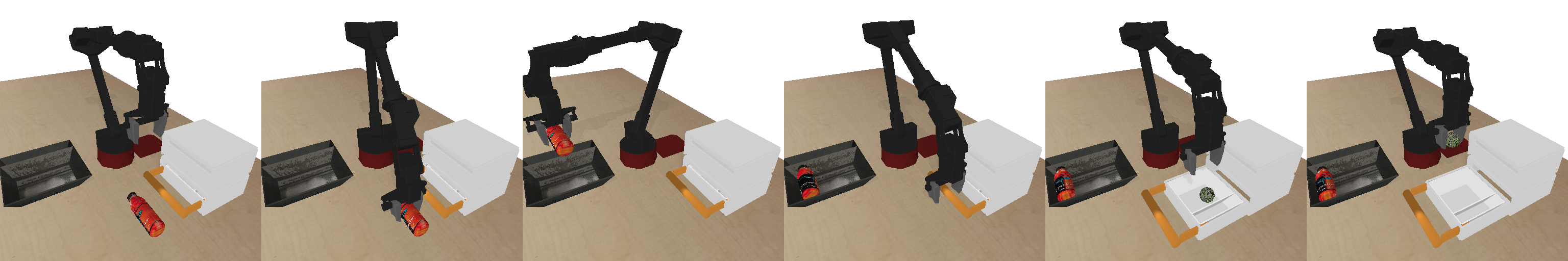}

    \caption{\footnotesize{\textbf{Grasping from the drawer with our learned policy.} The first row shows the training condition, which requires grasping from an open drawer. The robot only needs to grasp the object and take it out of the drawer to get a reward. The subsequent rows show the harder test-time initial conditions which require, respectively: opening the drawer before taking out the object, closing the top drawer before opening the bottom one and taking out the object, and removing an obstruction (red) bottle before opening the drawer. COG learns a policy that succeeds 70-75\% of the time for each type of initial condition, despite only having seen the object grasping from the open drawer.}}
    \vspace{-25pt}
    \label{fig:sim-task-drawer-grasp}
\end{center}
\end{wrapfigure}

We evaluate our approach in simulation (see Figures~\ref{fig:sim-task-pick-place} and~\ref{fig:sim-task-drawer-grasp}) and on a real-robot task with a WidowX low-cost arm (see Figure~\ref{fig:real-robot-task}).

\textbf{Pick and place.}
Our first simulated environment is shown in Figure~\ref{fig:sim-task-pick-place}. It consists of a 6-DoF WidowX robot in front of a tray containing a small object and a tray. The objective is to put the object inside the tray. The reward is +1 when the object has been placed in the box, and zero otherwise. A simple initial condition for this task involves the robot already holding the object at the start of the episode, while a harder initial condition is when the robot has to first pick up the object. For this simplified experimental setting, the prior data consists of 10K grasping attempts from a randomized scripted policy (that has a success rate of about 40\%, details of this policy are in Appendix~\ref{app:data_collection}). Note that we do not provide any labels for which attempts were successful, and which were not. The task-agnostic prior dataset also consists of behaviors that may be irrelevant for the task. The task-specific data consists of 5K placing attempts from a different scripted policy (with a high success rate of over 90\%, since the tray position is unchanged across trials), and these trajectories are labeled with rewards. Note that there is no single trajectory in our dataset that solves the complete pick and place task, but the prior and task-specific datasets have a non-zero overlap in their state distribution.

\textbf{Grasping from a drawer.} Our second and more complex simulated environment is shown in Figure~\ref{fig:sim-task-drawer-grasp}. 
It consists of a 6-DoF WidowX robot and a larger variety of objects. The robot can open or close a drawer, grasp objects from inside the drawer or on the table, and place them anywhere in the scene.
Some of these behaviors require various pre-conditions. For example, grasping an object from the drawer might require opening that drawer, which in turn might require moving an obstruction out of the way.
The task here consists of taking an object out of a drawer, as shown in Figure~\ref{fig:sim-task-drawer-grasp}. A reward of +1 is obtained when the object has been taken out, and zero otherwise. When learning the new task, the drawer always starts out open. The more difficult test conditions include ones where the drawer starts out closed, the top drawer starts out open (which blocks the handle for the lower drawer), and an object starts out in front of the closed drawer,
which must be moved out of the way before opening. These settings are illustrated in  Figure~\ref{fig:sim-task-drawer-grasp}. The prior data for this environment is collected from a collection of scripted randomized policies.
These policies are capable of opening and closing both drawers with 40-50\% success rates, can grasp objects in the scene with about a 70\% success rate, and place those objects at random places in the scene (with a slight bias for putting them in the tray).
The prior data does \emph{not} contain any interactions with the object inside the drawer and also contains data irrelevant to solving the task, such as behavior that blocks the drawer by placing objects in front of it. 
There are 1.5m datapoints (transitions) in the prior dataset, and 300K datapoints in the task-specific dataset.

\begin{table}[]

\small 

\begin{tabular}{l || r| r| r| r| r|| r}
\hline
\textbf{Task \& Initial Condition}  &  \textbf{No prior}  &  \multicolumn{3}{c|}{ \textbf{BC}}  & \textbf{SAC} & \textbf{COG (ours)}  \\
   & \textbf{data} & \textbf{init} & \textbf{all} & \textbf{oracle} &  & \\
\hline
\underline{place in box} & & & & & & \\ 
~~~object in gripper      &  \textbf{1.00 (0.00)}  & \textbf{1.00 (0.00)}   & 0.94 (0.01) & 0.95 (0.01) & 0.00 & \textbf{1.00 (0.00)} \\
~~~object in tray       &  0.00 (0.00)  & 0.00 (0.00)  &  0.02 (0.00) & 0.02 (0.01) & -- & \textbf{0.96 (0.04)}\\
\hline
\underline{grasp from drawer} & & & & & & \\
~~~open drawer   &  0.98 (0.01)  & \textbf{0.99 (0.00)}   &  0.63 (0.01) & 0.82 (0.01) & 0.00 & 0.98 (0.02) \\
~~~closed drawer   &   0.00 (0.00)  &  0.00 (0.00)  & 0.23 (0.01) &  0.27 (0.03) & -- & \textbf{0.68 (0.07)} \\
~~~blocked drawer 1 \!\!\! &  0.00 (0.00)  & 0.00 (0.00) & 0.34 (0.03) & 0.35 (0.03)  & -- & \textbf{0.78 (0.07)}\\
~~~blocked drawer 2 \!\!\! &   0.00 (0.00)  & 0.00 (0.00)  & 0.22 (0.03) & 0.25 (0.01) & -- & \textbf{0.76 (0.09)}\\
\hline
\end{tabular}

\caption{\footnotesize{\textbf{Results for simulated experiments.} Mean (Standard Deviation) success rate of the learned policies for our method (COG), its ablations and prior work. For the grasping from drawer task, blocked drawer 1 and 2 are initial conditions corresponding to the third and fourth rows of Figure~\ref{fig:sim-task-drawer-grasp}. Note that COG successfully performs both tasks in the majority of cases, from all initial conditions. SAC (--) diverged in our runs.}}
\vspace{-0.8cm}
\label{table:sim}
\end{table}

\textbf{Baselines and comparisons.} We compare COG to: \textbf{(1)} pre-training a policy via behavioral cloning on the prior data and then fine-tuning with offline RL on the new task, denoted as \textbf{BC-init}, \textbf{(2)} a na\"ive behavioral cloning baseline, denoted as \textbf{BC}, which trains with BC on all data, \textbf{(3)} an ``oracle'' version of behavioral cloning that is provided with handpicked successful trajectories on the new task, denoted as \textbf{BC-oracle} that is indicative of an upper bound on performance of selective cloning methods on a task, \textbf{(4)} a standard baseline off-policy RL method, \textbf{SAC}~\citep{haarnoja2018sacapps}, and finally \textbf{(5)} an ablation of our method without any prior data, indicated as \textbf{no prior data}.

\vspace{-7pt}
\subsection{Empirical Results} 
\vspace{-5pt}

\paragraph{Simulation results.} The results for our simulation experiments are summarized in Table~\ref{table:sim}. For all methods, we train the policy via offline RL until the success rate converges, and evaluate the final policy on 250 trials, after training has converged. We then average these success rates across three random seeds for each experiment. 
Detailed learning curves can be found in Appendix~\ref{app:exp_curves}.
We first note that our data-driven approach generally performs well for all initial conditions on both tasks. The policy is able to leverage the prior data to automatically determine that a closed drawer should be opened before grasping, and obstructions should be moved out of the way prior to drawer opening, despite never having seen complete episodes that involve both opening the drawer and taking out the object, and not having any reward or success labels in the prior data.

COG also significantly outperforms the behavioral cloning baselines, ablations, and a standard off-policy SAC algorithm for all novel initial conditions.
Since we use an entropy-regularized stochastic policy for the behavioral cloning variants (BC in Table~\ref{table:sim}), they are able to achieve a non-zero success rate, but the performance is far below what is achieved by COG. This is likely due to their inability to distinguish between behavior that supports the new task and behavior that is meaningful but not useful at the current time (for example, grasping an arbitrary object in the drawer task is not useful if that object is not blocking the drawer). Even when a behavioral cloning method is provided with handpicked successful trajectories for the new task, shown as BC-oracle in Table~\ref{table:sim}), we find that it is unable to complete the new task.
\begin{wrapfigure}{l}{0.43\textwidth}
\small
\vspace{-7pt}
\includegraphics[width=\linewidth]{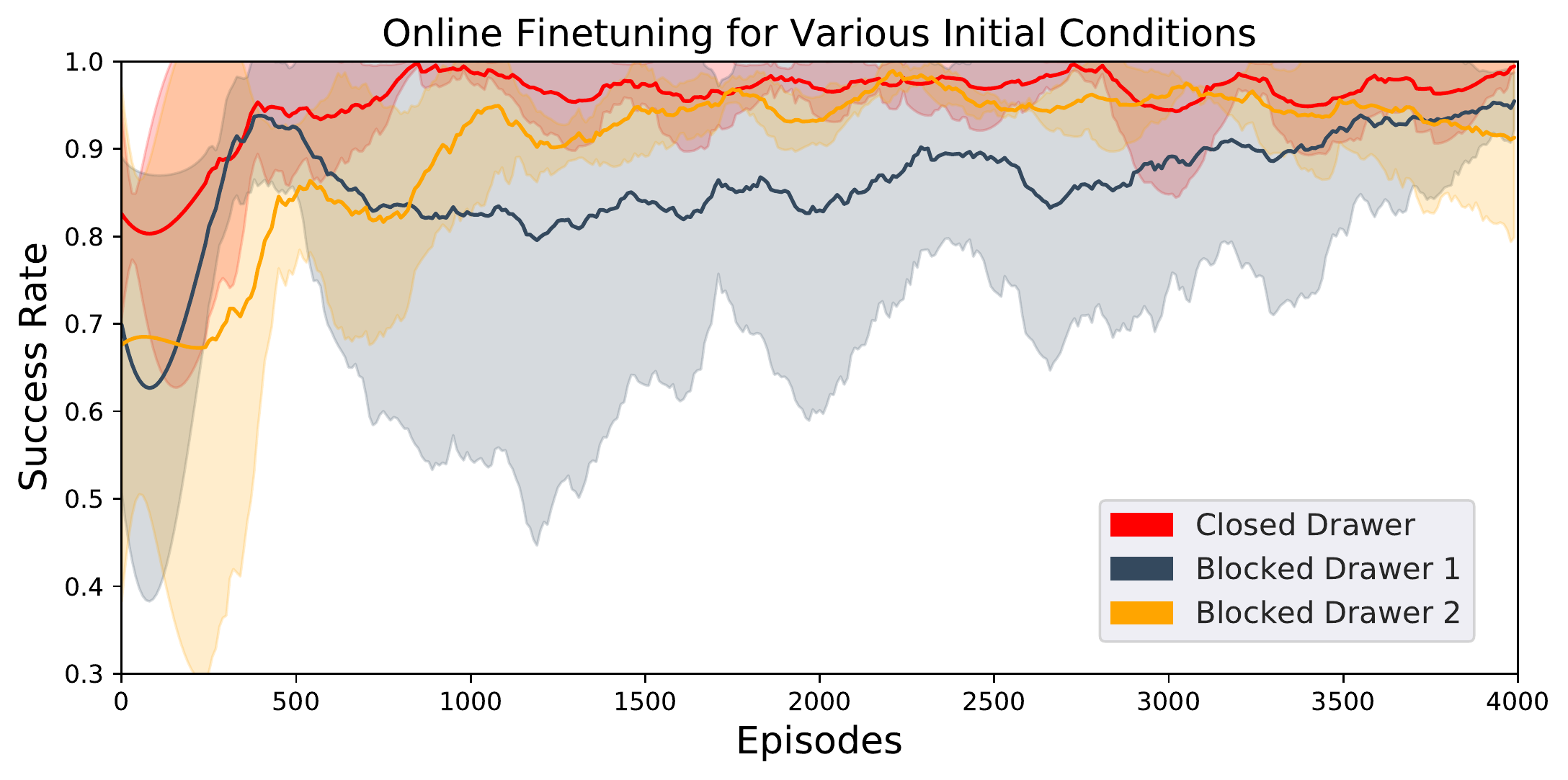}
\caption{\footnotesize{\textbf{Results from online fine-tuning.} We see that online fine-tuning  further improves the performance of the learned policy, bringing it to over 90\% success rates for all possible initial conditions for the drawer task, and only requires a small amount of additional data.}}
\vspace{-10pt}
    \label{fig:fine-tuning}
\end{wrapfigure}
Unsurprisingly, an ablation of our method that does not make use of prior data is only able to solve the task from states that were previously seen in the task-specific dataset. 
{BC-init}, which pre-trains the policy on the prior data via cloning followed by RL on the task data, is unable to solve the task for unseen initial conditions. We believe this is due to catastrophic forgetting of behavior learned during pre-training, and this adversely affects the policy's performance when evaluated on new initial conditions.

\textbf{Online fine-tuning.} While our method is able to obtain high success rates from offline learning alone, we also evaluated if the learned policies can be further improved via online fine-tuning. The results from these online fine-tuning experiments are shown in Figure~\ref{fig:fine-tuning}. We see that for all of the novel initial conditions for the drawer task, the policy is able to achieve a success rate of over 90\% from collecting only a small amount of additional episodes (500-4000, depending on the task). In our real world experimental setup (which we describe below), we are able to collect 3K episodes in a single day (autonomously), making this requirement quite feasible for real world problems. 
We compared this fine-tuning experiment against fine-tuning with SAC (starting from a behavior cloned policy), which performed substantially worse (see Figure~\ref{fig:sac-fine-tune-compare} in Appendix~\ref{app:sac-fine-tune-compare}), likely due to the low initial performance of the BC policy, and since the Q-function for SAC needs to be learned from scratch in this setting. Prior work has also observed that fine-tuning a behavior-cloned policy with SAC typically leads to some unlearning at the start, which further reduces the performance of this policy~\citep{nair2020accelerating}.
More details can be found in Appendix~\ref{app:sac-fine-tune-compare}.

\begin{wrapfigure}{r}{0.65\textwidth}
\small \begin{center}
\vspace{-16pt}
    \includegraphics[width=\linewidth]{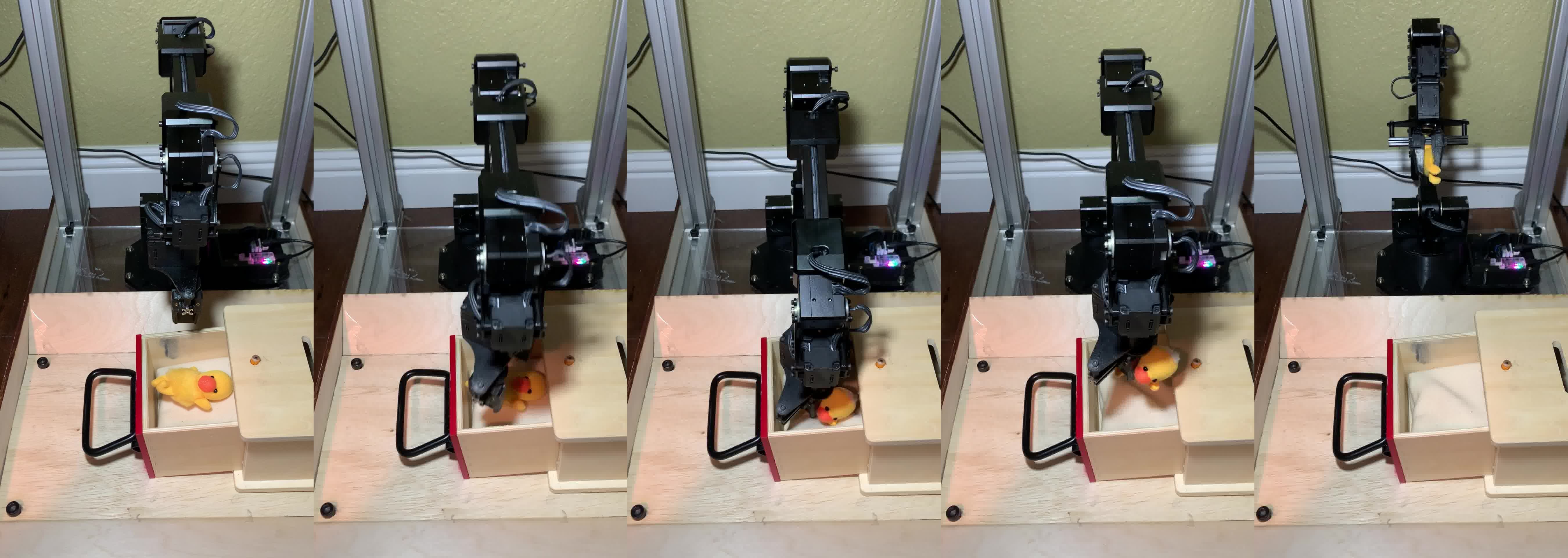}
    \vspace{-10pt}
    \includegraphics[width=\linewidth]{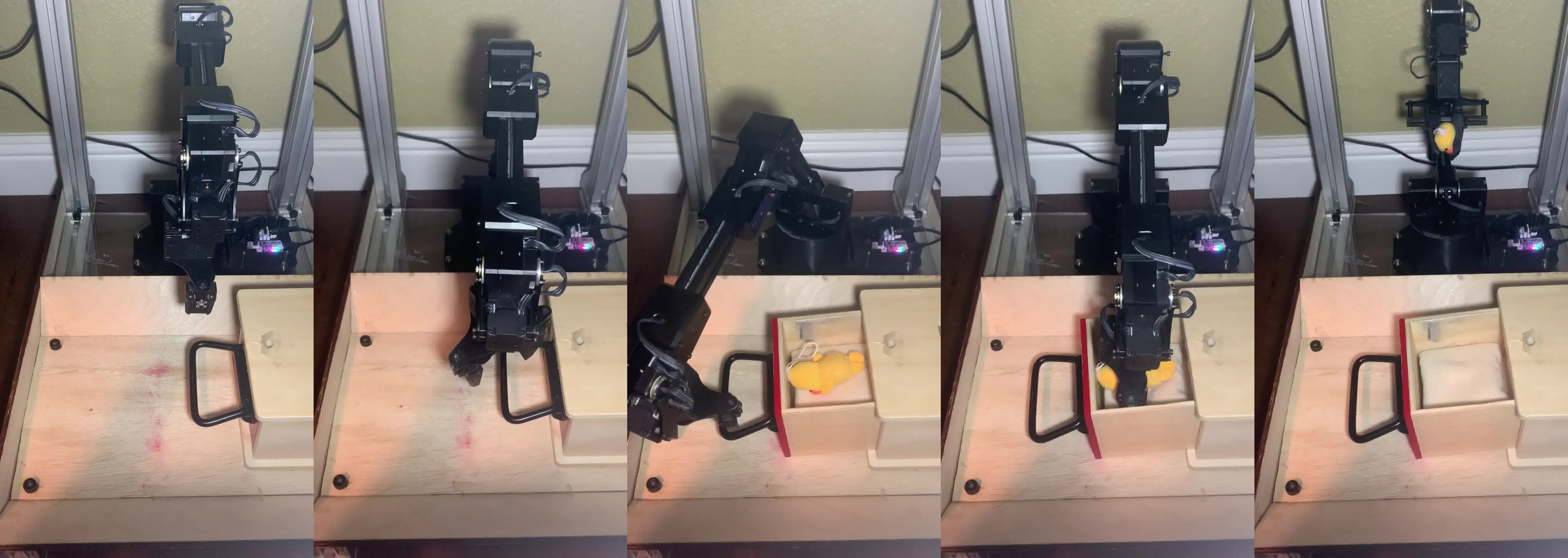}
    \caption{\footnotesize{\textbf{Real world drawer opening and grasping.} The top row shows the training condition, which requires grasping an object from an open drawer. The bottom row shows the behavior of the learned policy in the test condition, where the drawer starts closed, and shows a rollout from the learned policy, which never saw a complete trajectory of opening a drawer and grasping together at training time.}}
    \label{fig:real-robot-task}
\end{center}
\vspace{-20pt}
\end{wrapfigure}
\textbf{Real-world evaluation.} Our real world setup (see Figure~\ref{fig:real-robot-task}) consists of a WidowX robotic arm in front of a drawer, and an object inside the drawer. The task is to take the object out of the drawer. As before, the reward is +1 on completion, zero otherwise. During training for the new task, the drawer starts open. The prior dataset consists of drawer opening and closing from a scripted randomized policy (details in Appendix~\ref{app:data_collection}). As before, the prior dataset has no reward labels, and no instances of the new task (object grasping). The task-specific data is collected using a scripted grasping policy, which has a success rate of about 50\%. The prior dataset consists of 80K transitions collected over 20 hours, and the new task dataset has 80K transitions (4K grasp attempts) collected over 34 hours.
Our learned policy succeeds in \textbf{7/8} trials when the drawer starts out closed, and substantially outperforms the BC-oracle baseline, which \textbf{never} succeeds on this real-world task.

\section{Discussion}
We showed how model-free reinforcement learning can utilize prior data to improve the generalization of learned skills, effectively stitching together behaviors so that, when the initial conditions at test time differ from those under which a new task was learned, the agent can still perform the task by stitching together appropriate behaviors seen in the prior data. For instance, a robot that learned to grasp an object from an open drawer can automatically learn to open the drawer at test time when it starts closed, without ever having seen a complete opening and grasping trajectory. This sort of composition of skills is typically seen as the domain of model-based planning, and often approached in learning-based control from the standpoint of hierarchy. Our approach suggests that such seemingly complex multi-stage behaviors could emerge automatically as a consequence of model-free reinforcement learning with the right training data. In effect, we show that model-free RL can use prior data to learn the ``mechanics'' of the world, without any explicit hierarchy or model representation.
Our initial instantiation of this idea does leave considerable room for future research. First, we focus entirely on the offline setting, and performing online fine-tuning, both for the new task in isolation and for the multi-stage behaviors, would likely lead to better results. COG also employs a very na\"{i}ve approach for labeling rewards in the prior data, and more sophisticated reward inference methods could also improve performance. More broadly, we believe that further exploration into the ways that model-free RL methods can enable acquisition of implicit mechanical models of the world is a promising direction for data-driven robotics.

\section*{Acknowledgements}
We thank members of the Robotics and AI Lab (RAIL) at UC Berkeley for providing feedback on early drafts of this paper.
We also thank the anonymous reviewers for providing useful comments on the submitted version of this paper. 
This research was supported by the Office of Naval Research, the DARPA Assured Autonomy program, NSF IIS-1651843, and Berkeley DeepDrive, with compute support provided by Amazon, and Google. 




{\setstretch{0.7} \linespread{0.6}
    \bibliography{references}
}
\appendix

\newpage

\part*{Appendices}

\section{Experimental Setup Details}
\label{app:exp_details}

Both our real and simulated environments use the following 8-dimensional control scheme:

\begin{center} $\texttt{[x,y,z,alpha,beta,gamma,gripperOpen,moveToNeutral]}$
\end{center}

where the $\texttt{x,y,z}$ dimensions command changes in the end-effector's position in 3D space, $\texttt{alpha,beta,gamma}$ command changes in the end-effector's orientation, $\texttt{gripperOpen}$ is a continuous value from $[-1, 1]$ that triggers the gripper to close completely when less than $-0.5$ and open completely when greater than $0.5$, and $\texttt{moveToNeutral}$ is also a continuous value from $[-1, 1]$ that triggers the robot to move to its starting joint position when greater than $0.5$. The code for our environments can be found on our project website: \url{https://sites.google.com/view/cog-rl}. 

\subsection{Data Collection Policies}

\label{app:data_collection}
We describe our scripted data collection policies in this section. More details can be found in Algorithms 1-3. 

\paragraph{Scripted grasping.} Our scripted grasping policy is supplied with the object's (approximate) coordinates. In simulation, this information is readily available, while in the real world we use background subtraction and a calibrated monocular camera to approximately localize the object. Note that this information does not need to be perfect, as we add a significant amount of noise to the scripted policy's action at each timestep. After the object has been localized, the scripted policy takes actions that move the gripper toward the object (i.e action $\leftarrow$ object\_position $-$ gripper\_position). Once the gripper is within some pre-specified distance of the object, it closes the gripper (which is a discrete action). Note that this distance threshold is also randomized -- sampled from a Gaussian distribution with a mean of 0.04 and a standard deviation of 0.01 (in meters). It then raised the object until it is above a certain height threshold.  For the simulated pick and place environment, the scripted policy for grasping obtains a success rate of 50\%, while the success rate is 70\% for the drawer environment. For the real world drawer environment, the scripted success rate is 30\%. 

\paragraph{Scripted pick and place.} The pick part of the pick and place scripted policy is identical to the grasping policy described above. After the grasp has been attempted, the scripted policy uniformly randomly selects a point in the workspace to place the object on, and then takes actions to move the gripper above that point. Once within a pre-specified (and randomized) distance to that point, it opens the gripper. The policy is biased to sample more drop points that lie inside the tray to ensure we see enough successful pick and place attempts. Once the object has been dropped, the robot returns to its starting configuration (using the moveToNeutral action). 

\paragraph{Scripted place.} This policy is used in scenes where the robot is already holding the object at the start of the episode. The placing policy is identical to the place component of the pick and place policy described above.

\paragraph{Drawer opening and closing.} The scripted drawer opening policy moves the gripper to grasp the drawer handle, then pulls on it to open the drawer. The drawer closing policy is similar, except it pushes on the drawer instead of pulling it. Even if the correct action for a particular task might involve only opening the drawer, we collect data (without reward labels) that involves both opening and closing the drawer during  prior data collection. Gaussian noise is added to the policy actions at every timestep. After the opening/closing is completed, the robot returns to its starting configuration. 

\paragraph{Ending scripted trajectories with return to starting configuration} 
We ended the scripted trajectories with a return to the robot's starting configuration.  
We believe that this return to starting configuration increases the state-distribution overlap of the various datasets collected from scripted policies, making it possible to stitch together relevant trajectories from the prior dataset to extend the skill learned for the downstream task.

\begin{minipage}[t]{0.5\linewidth}
    \begin{algorithm}[H]
        \footnotesize
        \caption{Scripted Grasping}
        \label{alg:Scripted Grasping Policy}
        \begin{algorithmic}[1]
        \STATE $\text{threshold} \sim \mathcal N (0.04, 0.01)$
        \STATE numTimesteps $\leftarrow$ 30
        \FOR {t \textbf{in} (0, numTimesteps)}
            \STATE objPos $\leftarrow$ object position
            \STATE eePos $\leftarrow$ end effector position
            \STATE objGripperDist $\leftarrow$ distance(objPos, eePos)
            \IF {objGripperDist $>$ threshold}
                \STATE action $\leftarrow$ objPos $-$ eePos
            \ELSIF {gripperOpened}
                \STATE action $\leftarrow$ close gripper
            \ELSIF {object not raised high enough}
                \STATE action $\leftarrow$ lift upward
            \ELSE 
                \STATE action $\leftarrow$ 0
            \ENDIF
            \STATE noise $\sim \mathcal N(0, 0.2)$
            \STATE action $\leftarrow$ action + noise
            \STATE $(s, r, s') \leftarrow$ env.step(action)
        \ENDFOR
        \STATE
        \break
        \break
        \end{algorithmic}
    \end{algorithm}
\end{minipage}
\begin{minipage}[t]{0.5\linewidth}
    \begin{algorithm}[H]
        \footnotesize
        \caption{Scripted Pick and Place}
        \label{alg:Scripted Pick and Place}
        \begin{algorithmic}[1]
        \STATE $\text{threshold, dropDistThreshold} \sim \mathcal N (0.04, 0.01)$
        \STATE numTimesteps $\leftarrow$ 30
        \FOR {t \textbf{in} (0, numTimesteps)}
            \STATE eePos $\leftarrow$ end effector position
            \STATE dropPos $\leftarrow \begin{cases} \text{point above box} & \text{w/ prob. 0.5} \\ \text{point outside box} & \text{w/ prob. 0.5} \end{cases}$
            \STATE objectDropDist $\leftarrow$ distance(eePos, dropPos)
            \IF {object not grasped \textbf{AND} objectDropDist $>$ dropDistThreshold}
                \STATE Execute grasp using Algorithm \ref{alg:Scripted Grasping Policy}
            \ELSIF {objectDropDist $>$ boxDistThreshold}
                \STATE action $\leftarrow$ dropPos $-$ eePos
                \STATE action $\leftarrow$ lift upward
            \ELSIF {object not dropped}
                \STATE action $\leftarrow$ open gripper
            \ELSE
                \STATE action $\leftarrow$ 0
            \ENDIF
            \STATE noise $\sim \mathcal N(0, 0.2)$
            \STATE action $\leftarrow$ action + noise
            \STATE $(s, r, s') \leftarrow$ env.step(action)
        \ENDFOR
        \end{algorithmic}
    \end{algorithm}
\end{minipage}
\begin{minipage}[t]{0.5\linewidth}
    \begin{algorithm}[H]
        \footnotesize
        \caption{Scripted Drawer Opening/Closing}
        \label{alg:Scripted Drawer Opening}
        \begin{algorithmic}[1]
        \STATE $\text{threshold} \sim \mathcal N (0.04, 0.01)$
        \STATE numTimesteps $\leftarrow$ 30
        \FOR {t \textbf{in} (0, numTimesteps)}
            \STATE handlePos $\leftarrow$ handle center position
            \STATE eePos $\leftarrow$ end effector position
            \STATE targetGripperDist $\leftarrow$ dist(targetPos, eePos)
            \IF {targetGripperDist $>$ threshold \textbf{AND} not drawerOpened}
                \STATE action $\leftarrow$ targetPos $-$ eePos
            \ELSIF {not drawerOpened (or closed)}
                \STATE action $\leftarrow$ move left to open drawer, or right to close drawer
            \ELSIF {gripper not above drawer}
                \STATE action $\leftarrow$ lift upward
            \ELSE
                \STATE action $\leftarrow \texttt{moveToNeutral}$ 
                \STATE End scripted trajectory
            \ENDIF
            \STATE noise $\sim \mathcal N(0, 0.2)$
            \STATE action $\leftarrow$ action + noise
            \STATE $(s, r, s') \leftarrow$ env.step(action)
        \ENDFOR
        \end{algorithmic}
    \end{algorithm}
\end{minipage}
\begin{minipage}[t]{0.5\linewidth}
\end{minipage}

\newpage
\subsection{Neural Network Architectures}
\label{app:neural_net_arch}

\begin{figure}[h]
    \begin{center}
    \includegraphics[width=\linewidth]{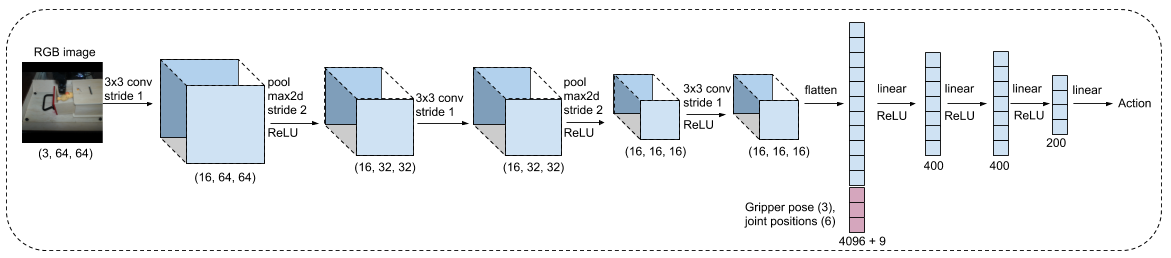}
    \caption{\footnotesize{\textbf{Neural network architecture for real robot experiments.} Here we show the architecture for the policy network for real robot experiments. The Q-function architecture is identical, except it also has the action as an input that is passed in after the flattening step. We map high dimensional image observations to low level robot commands, i.e. desired position of the end-effector, and gripper opening/closing.}}
    \label{fig:nn-real}
    \end{center}
\end{figure}

\begin{figure}[h]
    \begin{center}
    \includegraphics[width=\linewidth]{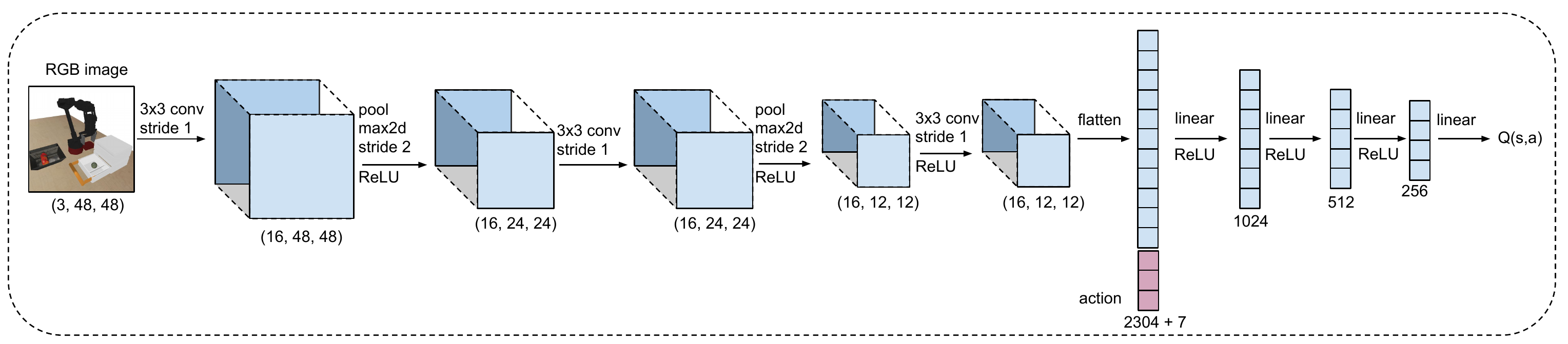}
    \caption{\footnotesize{\textbf{Neural network architecture for simulated experiments}. Here we show the architecture for the Q-function in our simulated experiments. The policy architecture is identical, except no action is passed to the network. Note that we omit the information about the gripper position and orientation, since including this information did not seem to make a difference in our simulated experiments.}}
    \label{fig:nn-sim}
    \end{center}
\end{figure}

Figures~\ref{fig:nn-real} and~\ref{fig:nn-sim} show the neural network architectures used in our real world and simulated experiments, respectively. We experimented with several different architectures (varying the number of convolutional layers from 2 to 4, and varying the number of filters in each layer from 4 to 16), and found these two architectures to perform the best. 



\subsection{Hyperparameters for Reinforcement Learning}
We used the conservative Q-learning (CQL)~\citep{kumar2020conservative} algorithm in our method.
Source code can be found on our project website: \url{https://sites.google.com/view/cog-rl}
We now present the hyperparameters used by our method below:

\begin{itemize}[leftmargin=*]
    \item \textbf{Discount factor}: 0.99 (identical to SAC, CQL),
    \item \textbf{Learning rates}: Q-function: 3e-4, Policy: 3e-5 (identical to CQL),
    \item \textbf{Batch size}: 256 (identical to SAC, CQL),
    \item \textbf{Target network update rate}: 0.005 (identical to SAC, CQL),
    \item \textbf{Ratio of policy to Q-function updates}: 1:1 (identical to SAC, CQL),
    \item \textbf{Number of Q-functions}: 2 Q-functions, $\min(Q_1, Q_2)$ used for Q-function backup and policy update (identical to SAC, CQL),
    \item \textbf{Automatic entropy tuning}: True, with target entropy set to $- \log |\mathcal{A}|$ (identical to SAC, CQL),
    \item \textbf{CQL version}: CQL($\mathcal{H}$) (note that this doesn't contain an additional $-\alpha \log \pi(\ba|\bs)$ term in the Q-function backup),
    \item \textbf{$\alpha$ in CQL}: 1.0 (we used the non-Lagrange version of CQL($\mathcal{H}$)),
    \item \textbf{Number of negative samples used for estimating logsumexp:} 1 (instead of the default of 10 used in CQL; reduces training wall-clock time substantially when learning from image observations)
    \item \textbf{Initial BC warmstart period}: 10K gradient steps
    \item \textbf{Evaluation maximum trajectory length}: 80 timesteps for simulated drawer environment, 40 timesteps for simulated pick and place. For real world drawer environment, this value is equal to 35 timesteps. 
    
\end{itemize}

\section{Comparison to BC + SAC for online fine-tuning}
\begin{figure}[h]
\centering
\includegraphics[width=0.5\linewidth]{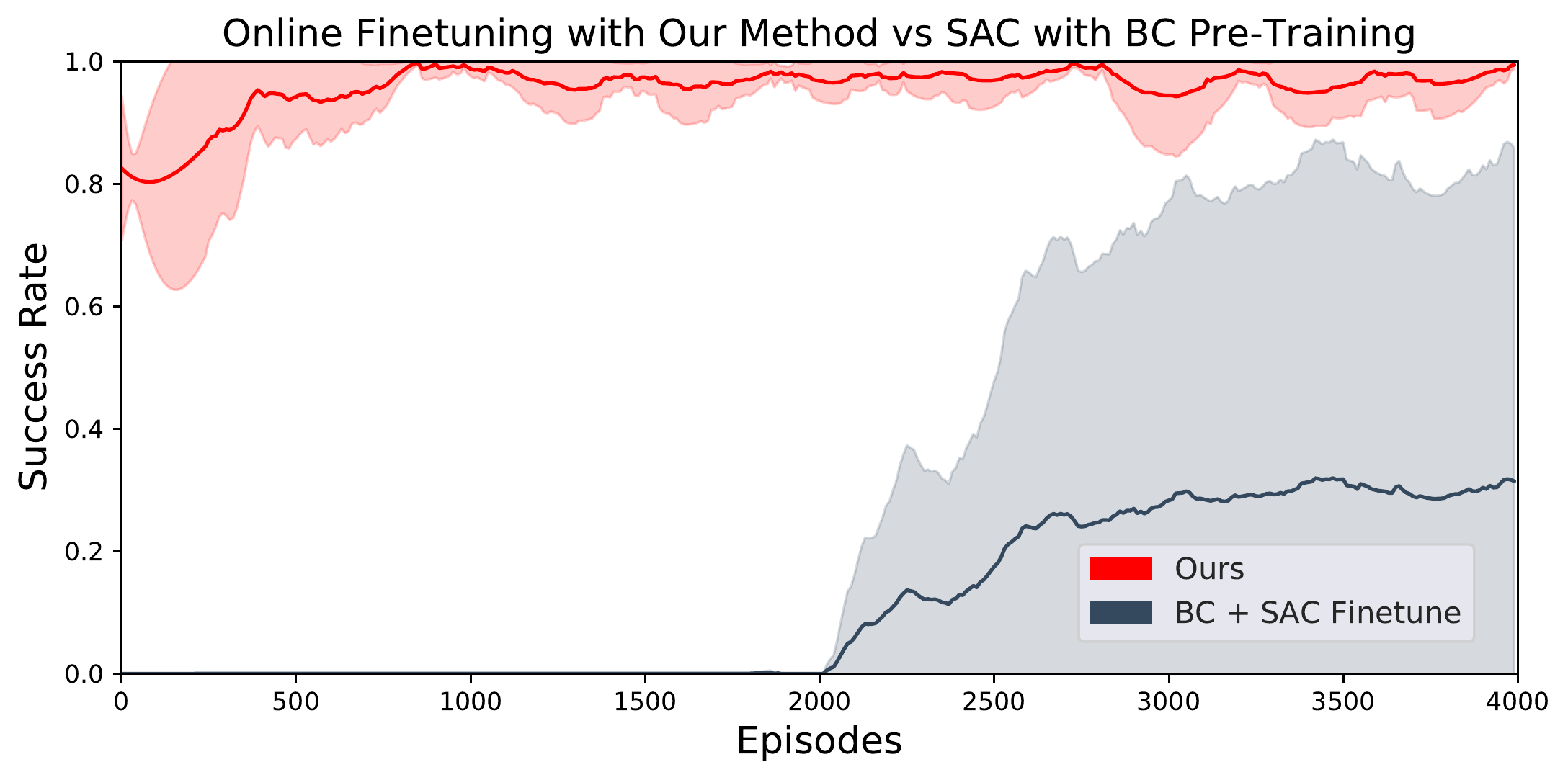}
\caption{\footnotesize{\textbf{Fine-tuning with CQL vs BC+SAC} We compared fine-tuning with CQL to fine-tuning a BC policy with SAC. SAC experiences some unlearning at the start (resulting in a success rate of zero at the start of training), and needs to collect a somewhat large number of samples before it can recover. Further, the variance across three random seeds was quite high for BC+SAC.}}
\label{fig:sac-fine-tune-compare}
\end{figure}

We compared our CQL fine-tuning results to fine-tuning a behavior-cloned policy with SAC, and observed that fine-tuning with CQL yields substantially better results. The comparison between between CQL fine-tuning, and this BC+SAC baselines is shown in Figure~\ref{fig:sac-fine-tune-compare} for the grasping from a drawer task (see Figure~\ref{fig:sim-task-drawer-grasp}), for the initial condition where the drawer starts out closed. We see that the initial SAC performance is low, which is partly due to the low success rate of the BC policy, and also because SAC typically undergoes some unlearning at the start of the fine-tuning process. This unlearning when fine-tuning with SAC has been observed in prior work~\citep{nair2020accelerating}, and is due to the fact that a randomly initialized critic is used to update the policy. For harder (i.e. long-horizon) tasks with more complicated initial conditions (such as blocked drawer 1 and blocked drawer 2), we were unable to get SAC to perform well from a BC initialization, even after we collecting over 5K new episodes. 
\label{app:sac-fine-tune-compare}

\newpage
\section{Learning Curves}
\label{app:exp_curves}

\begin{figure}[!htb]
\includegraphics[width=\linewidth]{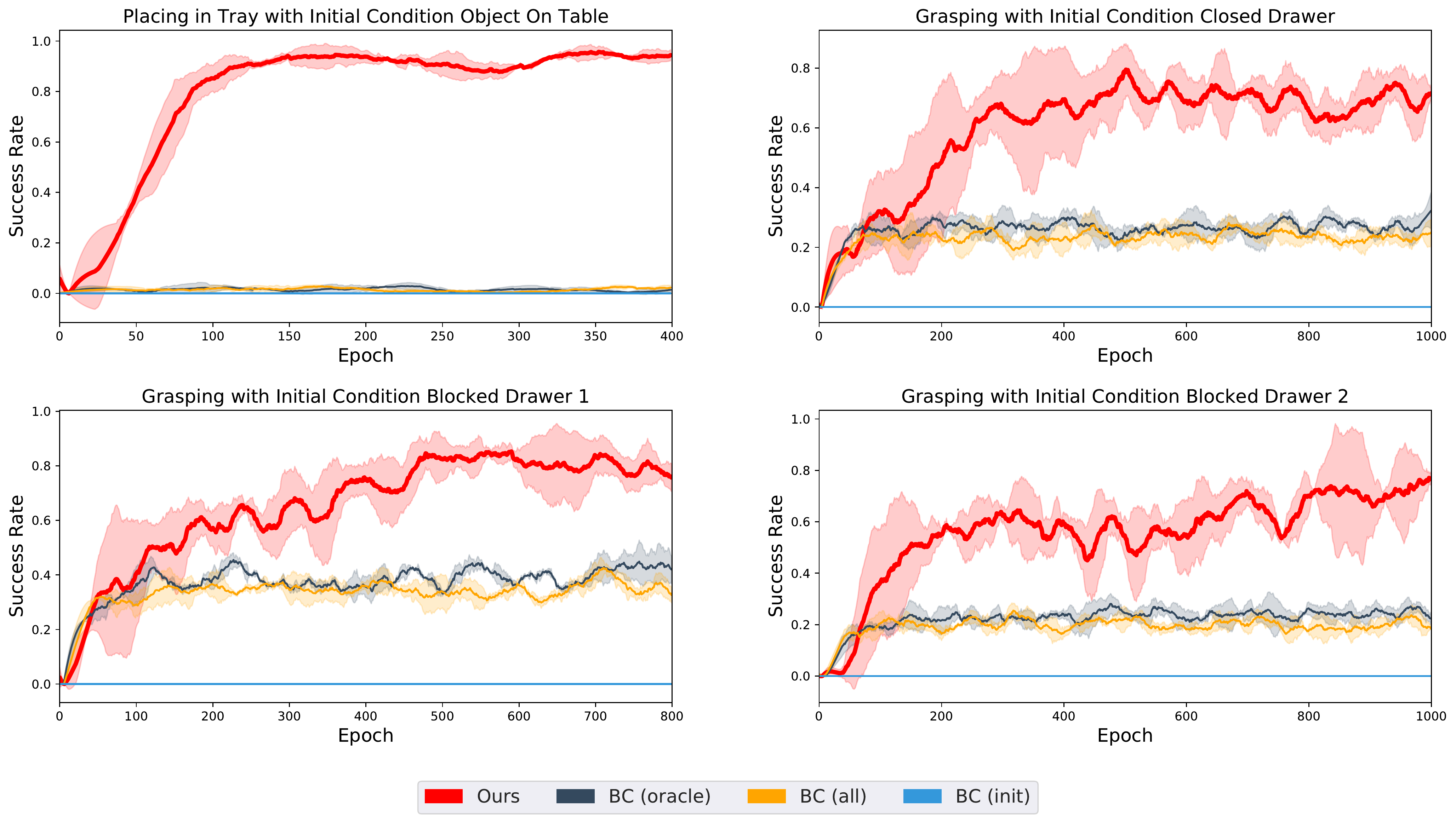}
\caption{\footnotesize{\textbf{Learning curves for simulated experiments by method and initial condition}. Here we compare the success rate curves of our method (COG) to the three behavioral cloning baselines in the four settings of Table \ref{table:sim} where prior data is essential for solving the task: the place in tray task with the object starting in the tray (upper left), as well as the grasp from drawer task with a closed drawer (upper right), blocked drawer 1 (lower left), and blocked drawer 2 (lower right). }}
\label{fig:learning-curves}
\end{figure}

Here are detailed learning curves for the experiments we summarized in Table 1. Note that the x-axis here denotes number of update steps made to the policy and Q-function, and not the amount of data available to the method. Since we operate in an offline reinforcement learning setting, all data is available to the methods at the start of training. We see that COG is able to achieve a high performance across all initial conditions for both the tasks. We substantially outperform comparisons to prior approaches that are based on pretraining using behavior cloning, including an oracle version that only uses trajectories with a high reward. 

\end{document}